\documentclass[preprint,12pt]{elsarticle}

\usepackage[table]{xcolor}
\usepackage{times}
\usepackage{latexsym}
\usepackage[T1]{fontenc}
\usepackage[utf8]{inputenc}
\usepackage{microtype}
\usepackage{inconsolata}

\usepackage{graphicx}
\usepackage{comment}
\usepackage{subcaption}
\usepackage{booktabs}
\usepackage{multirow}
\usepackage{pifont}
\usepackage{array}
\usepackage[mathscr]{eucal}
\usepackage{amsmath}
\usepackage{float}
\usepackage{amssymb}

\usepackage{tabularx}
\usepackage{adjustbox}

\usepackage[normalem]{ulem}
\useunder{\uline}{\ul}{}

\usepackage{color}

\newcommand{\xinbang}[1]{\textcolor{black}{\textbf{} #1}}
\newcommand{\devin}[1]{\textcolor{black}{\textbf{} #1}}

\journal{Knowledge-Based Systems}

\begin{document}

\begin{frontmatter}

\title{Large Language Models Can Better Understand Knowledge Graphs Than We Thought}

\author{Xinbang Dai\fnref{label1}}

\author{Yuncheng Hua\corref{cor1}\fnref{label2}}
\ead{devin.hua@unsw.edu.au}
\cortext[cor1]{Corresponding author.}

\author{Tongtong Wu\fnref{label3}}
\author{Yang Sheng\fnref{label4}}
\author{Qiu Ji\fnref{label4}}
\author{Guilin Qi\fnref{label1}}

\affiliation[label1]{organization={Southeast University},
            city={Nanjing},
            state={Jiangsu},
            country={China}}

\affiliation[label2]{organization={The University of New South Wales},
            city={Sydney},
            state={New South Wales},
            country={Australia}}

\affiliation[label3]{organization={Monash University},
            city={Melbourne},
            state={Victoria},
            country={Australia}}

\affiliation[label4]{organization={Nanjing University of Posts and Telecommunications},
            city={Nanjing},
            state={Jiangsu},
            country={China}}

\begin{abstract}
When we integrate factual knowledge from knowledge graphs (KGs) into large language models (LLMs) to enhance their performance, the cost of injection through training increases with the scale of the models.
Consequently, there is significant interest in developing prompt strategies that effectively incorporate KG information into LLMs.
However, the community has not yet comprehensively understood how LLMs process and interpret KG information in different input formats and organizations within prompts, and researchers often rely on trial and error.
To address this gap, we design extensive experiments to empirically study LLMs' comprehension of different KG prompts.
At the literal level, we reveal LLMs' preferences for various input formats (from linearized triples to fluent natural language text). At the attention distribution level, we discuss the underlying mechanisms driving these preferences.
% We then examine how the organization of structural knowledge affects LLMs and consider the robustness of LLMs to KGs for practical applications.
\devin{We then investigate how the organization of structured knowledge impacts LLMs and evaluate LLMs' robustness in processing and utilizing KG information in practical scenarios.}
Our experiments show that (1) linearized triples are more effective than fluent NL text in helping LLMs understand KG information and answer fact-intensive questions;
% (2) different LLMs exhibit varying preferences for various triple prompt organization formats;
\devin{(2) Different LLMs exhibit varying preferences for different organizational formats of triples;}
(3) LLMs with larger scales are more susceptible to noisy, incomplete subgraphs.

\end{abstract}

% %%Graphical abstract
% \begin{graphicalabstract}
% %\includegraphics{grabs}
% \end{graphicalabstract}

% %%Research highlights
% \begin{highlights}
% \item Research highlight 1
% \item Research highlight 2
% \end{highlights}

\end{frontmatter}

%% main text
%%

\section{Introduction}
\label{intro}

Recent studies commonly utilize databases containing extensive factual knowledge, such as knowledge graphs (KGs), to reduce hallucination in language models and enhance the quality of their generated content~\cite{pan2024unifying}. In the era of pre-trained language models (PLMs), integrating KG knowledge within the model during the training process has garnered significant interest within the community~\cite{sun2019ernie, xiong2019pretrained, su2021cokebert, arora2022metadata, chen2022dictbert}. However, as language models evolve, training large-scale language models (LLMs) with billions of parameters using KG data may encounter limitations, such as severe resource constraints or lack of public access to model architectures, training corpus, or training methods~\cite{ufuk2023role}. 
To address these issues, researchers increasingly focus on injecting external knowledge into LLMs through prompt engineering techniques~\cite{sorensen2022information, white2023prompt, li2023graph, wen2023mindmap}.
% \devin{This lightweight approach feeds knowledge from external KGs into LLMs in the form of prompts, demonstrating its effectiveness in addressing various challenges that rely solely on factual knowledge.}
This lightweight approach feeds knowledge from external KGs into LLMs in the form of prompts, demonstrating its effectiveness in addressing various challenges that rely solely on factual knowledge.
Some studies indicate that LLMs are highly sensitive to input patterns, and different input formats can impact model performance~\cite{sclar2023quantifying, voronov2024mind, zhan2024unveiling}.

\begin{figure}[htbp]
\centering
\includegraphics[width=0.9\columnwidth]{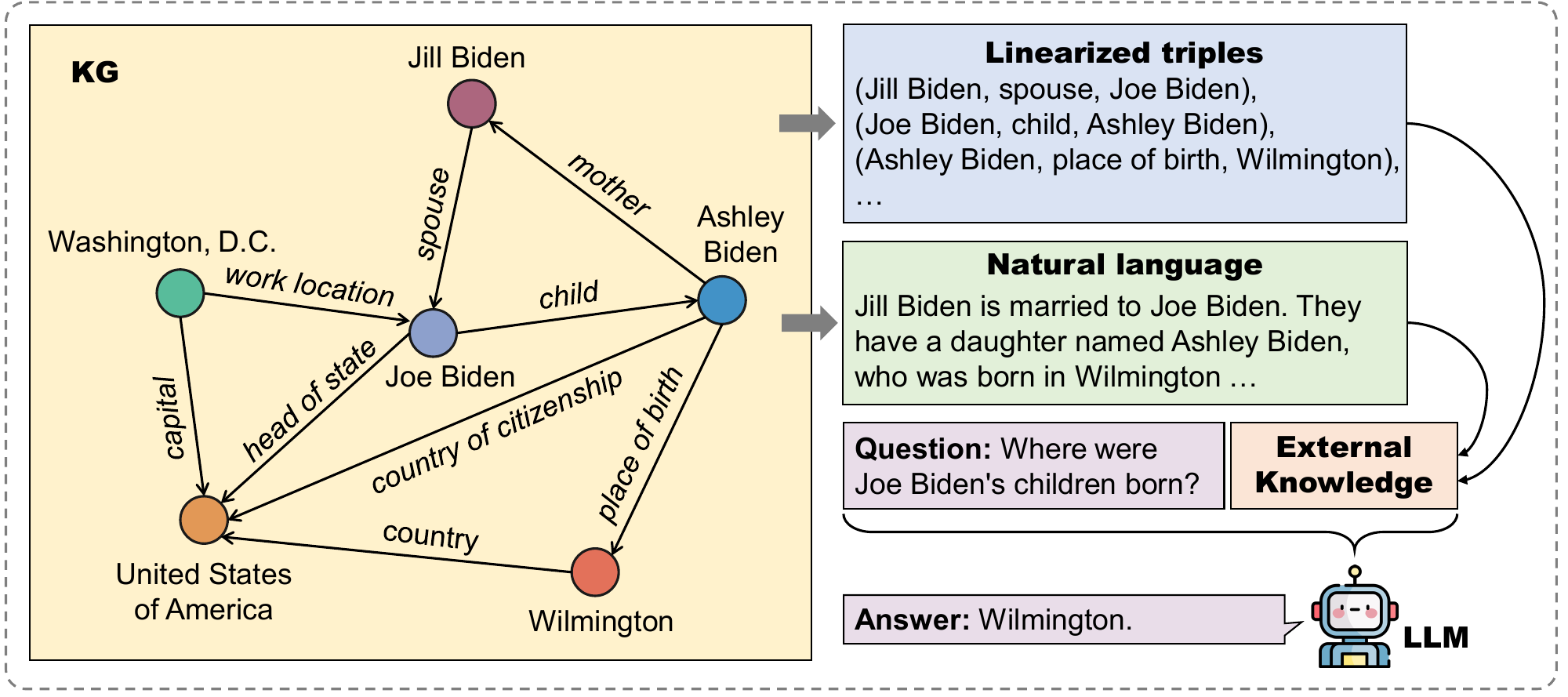}
\caption{KG is processed into different input formats to provide LLM with knowledge.}
\label{fig1}
\end{figure}

% As a highly structural knowledge, KG can be input into LLM in various formats. 
% \devin{As a highly structured form of knowledge, KGs can be integrated into LLMs in various ways.}
As a highly structured form of knowledge, KGs can be integrated into LLMs in various ways.
% As shown in Figure~\ref{fig1}, recent works process KGs into linearized triples and directly feed them into LLMs~\cite{Baek2023KnowledgeAugmentedLM, Sen2023KnowledgeGL}. 
% \devin{As illustrated in Figure~\ref{fig1}, recent studies have explored directly feeding linearized triples from KGs into LLMs~\cite{Baek2023KnowledgeAugmentedLM, Sen2023KnowledgeGL}.}
As illustrated in Figure~\ref{fig1}, recent studies have explored directly feeding linearized triples from KGs into LLMs~\cite{Baek2023KnowledgeAugmentedLM, Sen2023KnowledgeGL}.
Some other research employs KG-to-text generation approaches to convert structural knowledge prompts into natural language (NL) text, aiming to bridge the semantic gap between them~\cite{Ma2021OpenDQ, Xiong2022AutoQGSAF, wu2023retrieve, guo2023knowledgenavigator}.
However, generating text from KGs becomes a significant challenge when dealing with multi-relation subgraphs containing numerous triples (tens or even hundreds)~\cite{lai2024cognlg}. 
%
% Additionally, in some tasks where factual accuracy is crucial, the necessity of rewriting linearized triples into NL text to bridge the semantic gap has not been further explored.
%
Additionally, it is essential to recognize the necessity of bridging the semantic gap by rewriting linearized triples into NL text, as this process is complex and may be resource-intensive.
%
% Researchers currently rely on conjecture and experimentation.
%
The two types of approaches in Figure~\ref{fig1} highlight the community's limited understanding of how LLMs comprehend KGs. 
To further assist researchers in constructing more efficient KG-related prompt strategies, it is crucial to understand which KG input format is most beneficial for prompting LLMs in addressing fact-intensive queries.

% In this study, we aim to explore which KG input formats can better prompt LLMs in the background of prompt engineering~\cite{sahoo2024systematic}.
In this study, we aim to explore which KG input formats can better prompt LLMs~\cite{sahoo2024systematic}.
We curate our question-answering (QA) evaluation data based on multiple datasets for KG-related tasks (such as KGQA and relation extraction) and ask LLMs to answer complex questions based on different KG input formats. Compared to other KG-related tasks (such as link prediction and KG-to-text generation), the QA task is more complicated and offers a unique evaluation advantage. Our questions encompass entity enumeration, counting, ranking, comparison, and truthfulness assessment, thereby thoroughly evaluating LLMs' ability to adopt externally injected KG. LLMs may need to understand, retrieve, or associate external knowledge with their internal knowledge to better answer questions.
Despite the complexity of these questions, the QA evaluation method is simple but relatively objective. We can accurately evaluate the performance of LLMs by directly comparing the predicted answers with the golden answers.

Specifically, we first evaluate the preferences of LLMs for different KG input formats at both the literal and attention levels. The KG input formats range from unordered triples to fluent NL. At the literal level, we design two complementary experiments, \emph{Triple-to-Text} and \emph{Text-to-Triple}, to assess LLMs' preferences for KG input formats of varying scales and complexities. Compared to fluent NL text, we find that unordered linearized triples are more effective in assisting LLMs in answering knowledge-intensive questions. At the attention level, we analyze the attention distribution of LLMs across different input formats to investigate the reasons behind this phenomenon.
Furthermore, we investigate the impact of the organization of linearized triples on LLM performance by employing various prompt strategies. Finally, we evaluate the robustness of LLMs to noisy and incomplete external subgraphs. We hope our findings can provide the community with insights for better designing KG-related prompting strategies to enhance LLM performance. Our main findings are as follows:

\begin{itemize}

\item \textit{When using external knowledge to answer fact-intensive questions, LLMs prefer unordered structural data over fluent NL text.}

\item \textit{Sorting unordered linearized triples may not be necessary, as different LLMs prefer different prompt strategies. Researchers need to conduct meticulous experiments to develop more universal KG knowledge injection prompts.}

\item \textit{In the input format of linearized triples, LLMs with larger parameters are less robust to such inputs, i.e., they are more susceptible to noisy or incomplete KGs, resulting in performance degradation.}

\end{itemize}

\section{Related Works}
\label{sec:related_work}

\subsection{Injecting KG into LLM During Training.}
Injecting knowledge from KGs into LLMs during training has been extensively researched. This approach enables LLMs to grasp the semantics of KG embeddings through collaborative training~\cite{sun2019ernie, xiong2019pretrained, liu2020k, su2021cokebert, zha2022inductive, chen2022dictbert}. Although these methods have shown progress in smaller PLMs, their applicability to larger-scale LLMs  presents challenges that require careful consideration of model architecture, training methods, and other aspects. In addition, injecting knowledge within training may result in insufficient and incorrect internalized knowledge in LLMs~\cite{ji2023survey}.

\subsection{Integrating KG for LLM Using Prompt}
Integrating external structural knowledge into prompts to enhance LLM capabilities has become a common strategy. Some studies~\cite{Baek2023KnowledgeAugmentedLM, Sen2023KnowledgeGL} directly provide LLMs with linearized structural knowledge as part of the prompts. \cite{chen-2023-large} linearizes structural data into unified table rows, feeding them into the LLM to generate answers based on contextual examples. StructGPT~\cite{jiang2023structgpt} first selects entity and relation candidates, then uses multiple rounds of retrieving interface calls and LLMs as rankers to obtain answers.

However, other research suggests that inputs in NL format may be more model-friendly. These studies first convert structural knowledge into NL and then use it as prompts. UDT-QA~\cite{Ma2021OpenDQ} treats structural data as a form of knowledge expansion, converting it into NL text and adding it to the document repository for retrieval. \cite{Xiong2022AutoQGSAF} transforms subgraphs extracted from SPARQL queries into NL paragraphs, incorporating them into prompts to drive LLMs in generating NL questions. \cite{wu2023retrieve} believes that converting structural knowledge into high-quality NL text can substantially reduce the semantic gap between them. They employ KG-to-text generation models to rewrite structural knowledge and use it to help LLMs answer questions. KnowledgeNavigator~\cite{guo2023knowledgenavigator} performs efficient reasoning on KGs and uses templates to convert structural reasoning paths into NL, guiding LLM reasoning. 
\xinbang{These methods assume that NL formats are more suitable for LLMs, while neglecting the potential information noise and loss incurred during the format conversion process.}

\section{Unveiling LLMs’ Understanding of KG in Different Input Format}
\label{sec:3}

In this section, we conduct a comprehensive study to determine which input format of KGs is more beneficial for prompting LLMs. Our analysis is conducted on two levels: literal and attentional. At the literal level, we observe phenomena, while at the attention distribution level, we explore the underlying reasons.

% At the literal level (Section 3.1), we design two tasks: \emph{Triple-to-Text} and \emph{Text-to-Triple}. In the \emph{Triple-to-Text} task, we gradually convert the KG subgraph from the linearized triples into NL text generated by the model, which tests the LLM's understanding of KG across different scales and input formats. The experiments indicate that prompts composed of unordered triples are more effective for LLMs when answering fact-intensive questions. However, this raises a concern: is the decline in LLM performance on NL texts due to potential quality issues in the text generated by the model? To eliminate our doubt in \emph{Triple-to-Text}, we design a complementary task: \emph{Text-to-Triple}. Here, the NL text is manually written, and the corresponding subgraph is manually annotated, ensuring the quality of the KG as a prompt. Under the same evaluation metric, the experimental results from \emph{Text-to-Triple} align with the findings from \emph{Triple-to-Text}.

% \xinbang{At the literal level (Section 3.1), we design pipelines for prompting: \emph{Triple-to-Text} and \emph{Text-to-Triple}, which we evaluate using a QA task.}
\devin{At the literal level (Section~\ref{sec:3.1}), we design two pipelines for prompting LLMs: \emph{Triple-to-Text} and \emph{Text-to-Triple}, and use a QA task to evaluate these two pipelines.}
In \emph{Triple-to-Text}, we gradually convert the KG subgraph from the linearized triples into NL text generated by the model, which tests the LLM's understanding of KGs across different scales and input formats when answering questions. The experiments indicate that prompts composed of unordered triples are more effective for LLMs when answering fact-intensive questions. However, this raises a concern: is the decline in LLM performance on NL texts due to potential quality issues in the text generated by the model? 
\xinbang{To address concerns about \emph{Triple-to-Text}, we design a complementary pipeline: \emph{Text-to-Triple}.}
%
% Here, the NL text is manually written, and the corresponding subgraph is manually annotated, ensuring the quality of the KG as a prompt.
\devin{Here, the NL text is manually written, and the corresponding subgraph is manually annotated, ensuring the quality of the KG provided to the LLM as a prompt. Since both the NL text and the KG are already given, we do not need to rely on generative models to convert the KG into NL text. This allows for a more fair and controlled experimental comparison between the NL text and the KG.}
%

% At the attention level (Section 3.2), we capture the LLM's focus on KG knowledge within the prompt.
\devin{At the attention level (Section~\ref{sec:3.2}), we capture how the LLMs place greater focus on KG knowledge within the prompt.}
The experiments reveal that, regardless of whether a single KG format (linearized triples \textbf{or} NL text) or double formats (linearized triples \textbf{and} NL text) are provided, LLMs consistently demonstrate a greater capacity to capture answer information from linearized triples.

\subsection{Literal Level Evaluation of LLM's Understanding of KG}
\label{sec:3.1}

\subsubsection{LLM's Understanding of \emph{Triple-to-Text}}
\label{ssec:3.1.1}

% In terms of the \emph{Triple-to-Text} Pipeline, we analyze the understanding capabilities of LLMs regarding KG subgraphs from two aspects: \textbf{KG subgraph scale} and \textbf{KG input format}. 
\devin{In terms of the \emph{Triple-to-Text} pipeline, we analyze the LLM's ability to understand knowledge graph subgraphs from two perspectives: \textbf{KG subgraph scale} and \textbf{KG input format}.}
To quantitatively study the impact of the subgraph scale on LLMs, we propose \emph{Core Reasoning Path Generation} and \emph{Controlled Neighbour Node Search} to control the subgraph scale. Subsequently, the subgraphs are fed to the LLMs in various input formats to examine their preferences.

\begin{figure}[t]
\centering
\includegraphics[width=\columnwidth]{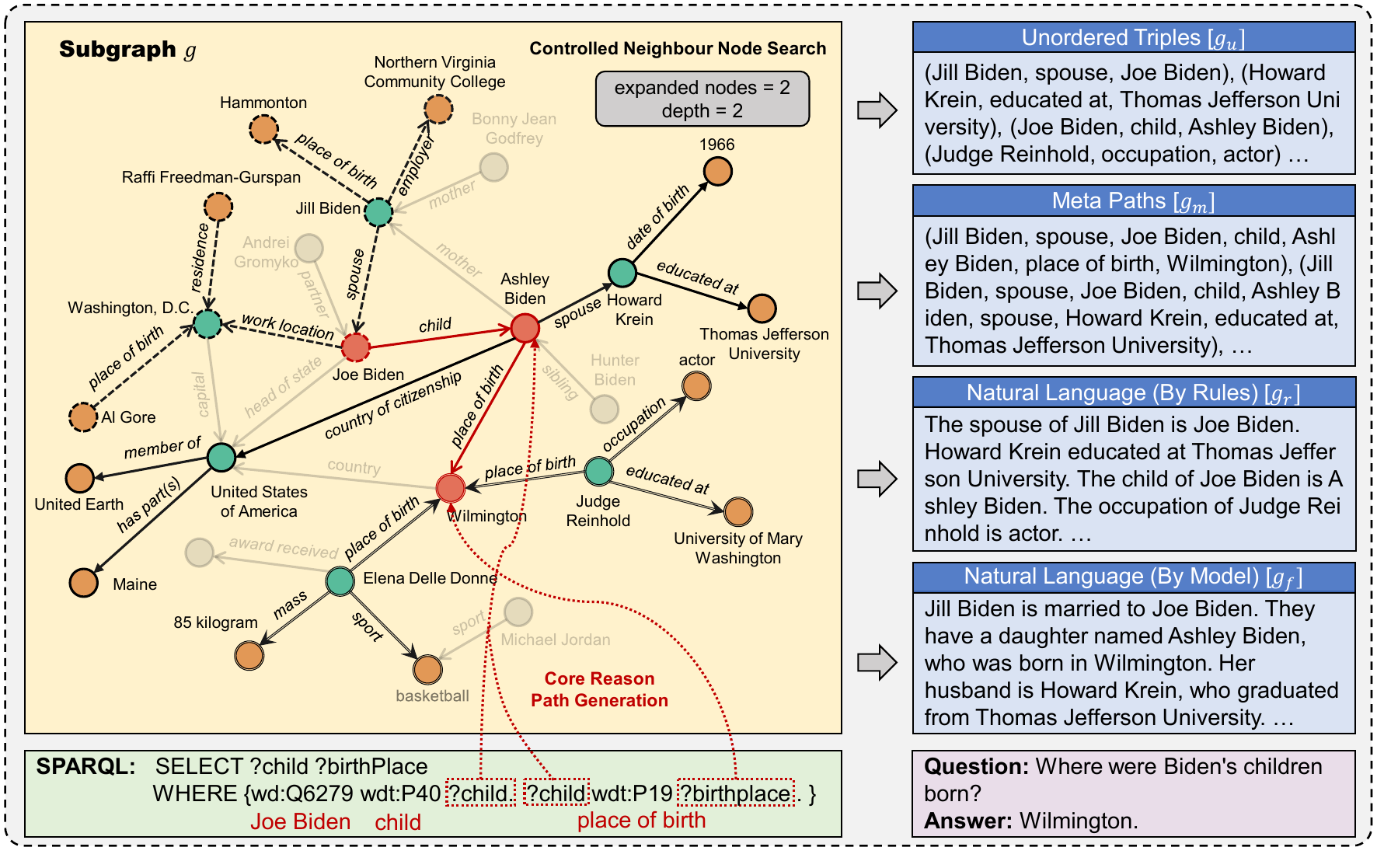}
\caption{There are six categories of our expansion method. (1) \textbf{\emph{expanded nodes} = 0, \emph{depth} = 0}: only providing core reasoning paths; (2) \textbf{\emph{expanded nodes} = 0.5, \emph{depth} = 1}: expanding each node on each core path by one neighbouring node, with a 50\% probability of deleting this expansion node; (3) \textbf{\emph{expanded nodes} = 1, \emph{depth} = 1}: expanding each node on each core path by one neighbouring node; (4) \textbf{\emph{expanded nodes} = 2, \emph{depth} = 1}: expanding each node on each core path by two neighbouring nodes; (5) \textbf{\emph{expanded nodes} = 1, \emph{depth} = 2}: starting from nodes on the core path, expanding to 2-hop neighbouring nodes, expanding one node at a time; (6) \textbf{\emph{expanded nodes} = 2, \emph{depth} = 2}: starting from nodes on the core path, expanding to 2-hop neighbouring nodes, expanding two nodes at a time (shown in this figure).}
\label{fig2}
\end{figure}

\paragraph{Core Reasoning Path Generation}
Firstly, to obtain the core reasoning path \( C \) for each question, we utilize the SPARQL queries\footnote{https://www.wikidata.org/wiki/Wikidata:SPARQL\_tutorial} provided in the KGQA dataset to retrieve all answers and their corresponding multiple reasoning paths. Specifically, we use the Wikidata endpoint to extract all constraint variable IDs from the SPARQL query for each question and then convert these IDs into their corresponding English labels. As shown in Figure~\ref{fig2}, by arranging these constraint variables in sequence, we derive the core reasoning path in the KG (in red). 

\paragraph{Controlled Neighbor Node Search}
Once the reasoning path \( C \) is obtained, we control the number of adjacent nodes and the number of rounds for subgraph \( g \) expansion using the parameters \emph{expanded nodes} and \emph{depth}. As illustrated in Figure~\ref{fig2}, when \emph{expanded nodes} = 2 and \emph{depth} = 2, two neighbouring nodes are randomly selected for expansion from each node on the reasoning path, and the expansion round is two. For example, \emph{Joe Biden} expands to \emph{Jill Biden} and \emph{Washington, D.C.}, while \emph{Ashley Biden} expands to \emph{United States of America} and \emph{Howard Krein}. The nodes from the first expansion round are shown in green, and those from the second round are in orange.
As shown in Figure~\ref{fig2}, the parameter \emph{expanded nodes} is selected as \{0.5, 1, 2\}, and \emph{depth} is selected as \{1, 2\}. Consequently, we construct six different subgraphs to evaluate the reasoning performance of the LLM.

\paragraph{KG Input Format}
To investigate the impact of different knowledge formats for LLMs, we refer to some intermediate steps in recent works and devise five levels of KG injection methods: (1) Omitting Subgraph: we ask LLMs to respond to the questions without subgraph \( g \), denoted as \( g_o \). (2) Unordered Triples: we randomly shuffle all triples in subgraph \( g \), denoted as \( g_u \). (3) Meta Path: we connect triples sharing the same head or tail entity to construct meta-paths~\cite{gao2020rdf}, denoted as \( g_m \). (4) Natural Language (rule-based): we use heuristic rules to convert triples into NL text~\cite{wang2022knowledge}, denoted as \( g_r \). (5) Natural Language (model-based): we employ a data-to-text generation model MVP~\cite{tang2022mvp} to convert \( g \) into fluent NL text~\cite{tang2022mvp}, denoted as \( g_f \).
Figure~\ref{fig2} illustrates all input formats. There are a total of 25 possible combinations of extension and injection modes. Omitting subgraphs \( g_o \) is considered one mode because it does not incorporate subgraph knowledge.

\paragraph{Datasets}
Wikidata~\cite{vrandevcic2014wikidata} is a large-scale, high-quality KG that is frequently updated. 
% We select three KGQA datasets based on Wikidata, which include SPARQL queries: QALD-7~\cite{usbeck20177th}, LC-QuAD 2.0~\cite{dubey2019lc}, and KQAPro~\cite{cao2020kqa}. 
% \devin{We selected three KGQA datasets based on Wikidata, i.e., QALD-7~\cite{usbeck20177th}, LC-QuAD 2.0~\cite{dubey2019lc}, and KQAPro~\cite{cao2020kqa}.
% All of these three datasets already provide ground-truth SPARQL query annotations.}
We selected three KGQA datasets based on Wikidata, i.e., QALD-7~\cite{usbeck20177th}, LC-QuAD 2.0~\cite{dubey2019lc}, and KQAPro~\cite{cao2020kqa}.
All of these three datasets already provide ground-truth SPARQL query annotations.
These datasets serve as the foundation for generating our datasets. QALD-7 contains 215 training questions and 50 test questions. LC-QuAD 2.0 comprises 24k training questions and 6046 test questions. KQAPro includes 94k training questions and 10k test questions. From these datasets, we use a SPARQL endpoint to retrieve answers from Wikidata, filtering out questions with incorrect or unanswerable results. We also delete the questions where the core reasoning path cannot be extended to two rounds, such as when the path contains numerical or other attribute information. After filtering, QALD-7 retains 64 questions. For LC-QuAD 2.0 and KQAPro, considering the cost of calling the API, we randomly select 2000 questions from each dataset.

\paragraph{Experimental Setup} We employ ChatGPT\footnote{https://openai.com/index/chatgpt},  Vicuna 7B and 13B~\cite{zheng2023judging} to evaluate the data, with all model parameters fixed. ChatGPT and Vicuna represent two mainstream series of LLMs, both demonstrating decent performance. To ensure a fair comparison, we use the exact match (EM) metric based on the standard SQuAD~\cite{rajpurkar2016squad} to evaluate the performance of the LLM. Following the approach in~\cite{tan2023evaluation}, we maintain an alias table for entities in our answers, derived from Wikidata, to match different mentions of the same entity in the answers.

\begin{figure}[t]
\centering
\includegraphics[width=\columnwidth]{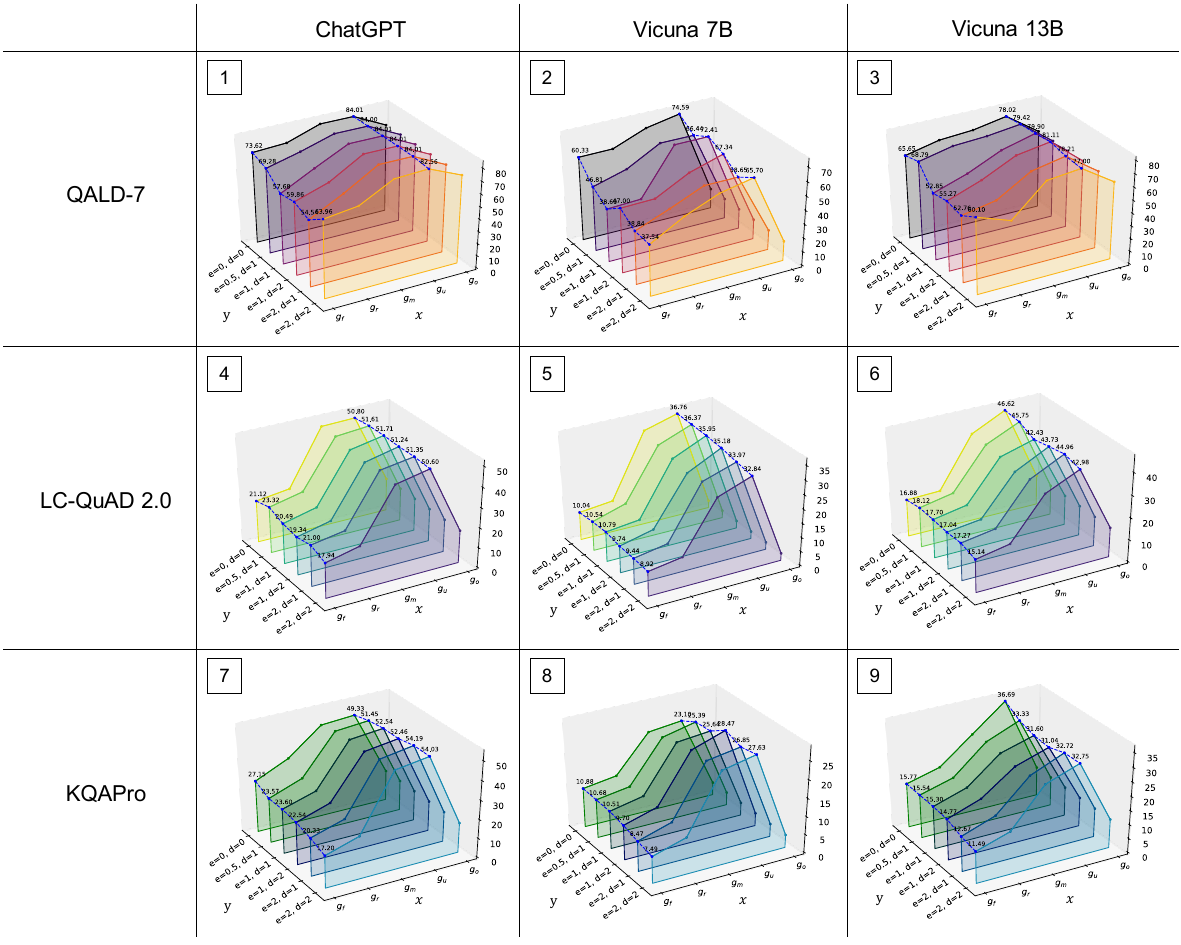}
\caption{Performance of LLM in Triple-to-Text \xinbang{Pipeline}. The x-axis represents various input formats, while the y-axis indicates different subgraph sizes. The parameter \( e \) denotes \emph{expanded nodes}, and \( d \) represents \emph{depth}.}
\label{fig3}
\end{figure}

\paragraph{Results analysis} Figure~\ref{fig3} shows that the unordered linearized triple input \( g_u \) consistently outperforms other methods. The limited size (only 64 questions) of the QALD-7 dataset initially obscures the advantages of our process. However, as the questions' complexity and the data's scale increase (as demonstrated in LC-QuAD 2.0 and KQAPro), the linearized triples demonstrate better performance in the knowledge prompt. Moreover, the downward trend from \( g_u \) to \( g_f \) on the x-axis is consistent across different models and datasets, indicating that unordered triple knowledge can help LLM better answer multi-hop fact-intensive questions.

When examining the impact of subgraph size on LLM performance along the y-axis, we observe that LLMs do not experience significant performance degradation when faced with larger subgraphs. The blue dashed lines indicate the performance variation of LLMs under different subgraph scales with the input format of \( g_u \) and \( g_f \). This decline is not as pronounced as the performance decline caused by changing input formats. 
%
% This indicates that LLMs are not modeling structural information from linearized triple prompts, making them insensitive to the scale and structure of subgraphs.
%
This indicates that changes in the scale and structure of subgraphs have a limited impact on the performance of LLMs, while the input format is a more significant factor in prompts affecting LLMs.
To further explore the preference of LLMs for linearized triple formats, in Section~\ref{sec:4.2}, we conduct experiments at the attention layer level. We observe that LLMs are more adept at focusing on knowledge relevant to the answers within linearized triples.

\subsubsection{LLM's Understanding of \emph{Text-to-Triple}}
\label{ssec:3.1.2}

\begin{figure}[htbp]
\centering
\includegraphics[width=\columnwidth]{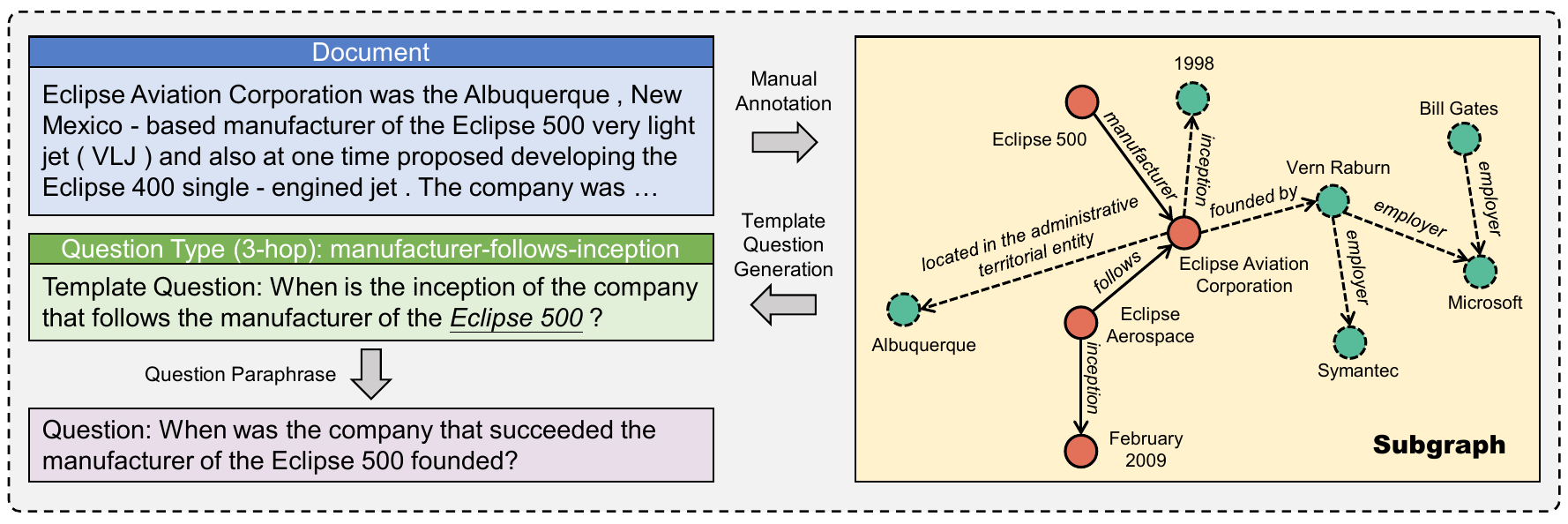}
\caption{We employ a human-annotated KG mapping with a document for generating multi-hop questions, then evaluate the performance of LLM in answering these fact-related questions with \emph{Documents} and \emph{Subgraph}.}
\label{fig4}
\end{figure}

\devin{In \emph{Text-to-Triple} pipeline, we use manually annotated NL text, rather than model-generated text, to evaluate LLMs' ability to understand external knowledge.}
In \emph{Triple-to-Text}, the quality of NL text \( g_f \) generated from KG may contain errors, potentially affecting the ability of LLMs to understand KGs in NL format.
In order to eliminate such concern, it is essential to provide LLMs with human-written NL text and establish a mapping from text to KG. 
%
% The document-level relation extraction dataset DocRED~\cite{yao2019docred} provides human-craft annotations for triples, which can be considered as ground-truth mappings of NL text. Therefore, we process the DocRED and generate a new QA dataset, addressing the challenge of ensuring NL text quality in triple generation.
%
% As illustrated in Figure~\ref{fig4}, all documents in DocRED are human-written, and all mapped triples are manually annotated and aligned with entities and relations in Wikidata. This dataset emphasizes cross-sentence reasoning, allowing the mapped triples within each document to form a complete, small-scale subgraph. We consider this subgraph as a complete structural representation of all entities and relations involved in the document. Based on these subgraphs, we can extract all reasoning paths of different hops to generate fact-related questions.
%
% \xinbang{The document-level relation extraction datasets, DocRED~\cite{yao2019docred}, which is based on Wikidata, and ChemDisGene~\cite{zhang2022distant}, which is based on Comparative Toxicogenomics Database\footnote{https://ctdbase.org} for biomedical data, provide human-craft annotations for triples which can be considered as ground-truth mappings of NL text.}
\devin{The document-level relation extraction datasets, DocRED~\cite{yao2019docred}, derived from Wikidata, and ChemDisGene~\cite{zhang2022distant}, based on the Comparative Toxicogenomics Database\footnote{https://ctdbase.org} for biomedical data, provide human-crafted annotations for triples, which serve as ground-truth mappings for NL text.}
\xinbang{We process these datasets and generate two new QA datasets, addressing the challenge of ensuring NL text quality in triple generation. 
We also utilize the QA dataset derived from ChemDisGene to examine LLMs' understanding capabilities when faced with different input formats of \textbf{domain-specific KG prompts}.
%
% Compared to common knowledge, LLMs often rely on user-provided external knowledge when handling domain-specific queries. Hence, it is crucial to understand LLMs' comprehension of domain-specific KGs in various input formats as external knowledge.
}

All documents in DocRED are human-written, and all mapped triples are manually annotated and aligned with entities and relations in Wikidata. \xinbang{The ChemDisGene document set is comprised of biomedical paper abstracts, with mapped triples manually curated by a team of biologists. In both datasets, the triples within each document form a complete, small-scale KG subgraph.} We consider this subgraph as a complete structural representation of all entities and relations involved in the document. As shown in Figure~\ref{fig4}, based on these subgraphs, we can extract all reasoning paths of different hops to generate fact-related questions.

\paragraph{Data Generation} To generate high-quality QA pairs from each document, we refer to the method used in MetaQA~\cite{zhang2018variational} for constructing multi-hop questions from triples. As shown in Figure~\ref{fig4}, we fill the entities into various manually crafted multi-hop templates to ensure semantic coherence. Then, we use ChatGPT to paraphrase these questions, enhancing their diversity.
DocRED consists of 5,053 Wikipedia documents, each associated with an annotated subgraph. We randomly select 800 documents that contain at least 3-hop paths and create 1-hop, 2-hop, and 3-hop questions to evaluate the LLM's ability based on subgraphs.
%
% \xinbang{ChemDisGene consists of 523 abstracts, each annotated by biologists. Based on the scale of corresponding subgraphs in the documents, we generate 518 1-hop questions, 491 2-hop questions, and 243 3-hop questions.}
\devin{ChemDisGene consists of 523 abstracts, each annotated with the corresponding triples by the biologists. Based on the scale of these subgraphs corresponding to the documents, we generate 518 1-hop questions, 491 2-hop questions, and 243 3-hop questions.}
We separately provide the unordered triples and human-written NL documents to the LLM to answer these questions.
% We present multi-hop questions constructed via documents in Appendix~\ref{appendix_1}.

\paragraph{Experimental Setup} To eliminate doubts about the quality of NL knowledge generated by the model, we use the same LLM configuration and evaluation metric as \emph{Triple-to-Text}, with the only difference being the replacement of \(g_f\) type knowledge with human-written documents.

\begin{table}[htbp]
\setlength{\tabcolsep}{3pt}
\resizebox{\textwidth}{!}{ 
\begin{tabular}{lcccccccccccc}
\toprule

 & \multicolumn{6}{c}{DocRED} & \multicolumn{6}{c}{\xinbang{ChemDisGene}} \\ 

\cmidrule(r){2-7} \cmidrule(r){8-13}
 
 & \multicolumn{2}{c}{1 hop} & \multicolumn{2}{c}{2 hop} & \multicolumn{2}{c}{3 hop} & \multicolumn{2}{c}{1 hop} & \multicolumn{2}{c}{2 hop} & \multicolumn{2}{c}{3 hop} \\ 

\cmidrule(r){2-3} \cmidrule(r){4-5}  \cmidrule(r){6-7} \cmidrule(r){8-9} \cmidrule(r){10-11} \cmidrule(r){12-13}
 
LLMs & Text & Triple & Text & Triple & Text & Triple & \multicolumn{1}{c}{Text} & \multicolumn{1}{c}{Triple} & \multicolumn{1}{c}{Text} & \multicolumn{1}{c}{Triple} & \multicolumn{1}{c}{Text} & \multicolumn{1}{c}{Triple} \\ 

\cmidrule(r){1-1} \cmidrule(r){2-3} \cmidrule(r){4-5}  \cmidrule(r){6-7} \cmidrule(r){8-9} \cmidrule(r){10-11} \cmidrule(r){12-13}

ChatGPT & 55.25 & \textbf{73.38} & 14.25 & \textbf{19.88} & 14.00 & \textbf{18.25} & 43.47 & \textbf{57.54} & 7.05 & \textbf{9.80} & 5.70 & \textbf{7.57} \\
Vicuna 7B & 34.88 & \textbf{50.13} & 9.50 & \textbf{11.00} & 8.63 & \textbf{10.50} & 25.21 & \textbf{32.53} & 3.49 & \textbf{7.28} & 2.72 & \textbf{4.91} \\
Vicuna 13B & 47.62 & \textbf{73.26} & 15.37 & \textbf{16.38} & 13.87 & \textbf{14.75} & 38.05 & \textbf{45.55} & 5.10 & \textbf{8.33} & 3.74 & \textbf{5.04} \\
GPT-4 & 59.16 & \textbf{75.48} & 19.37 & \textbf{22.96} & 17.68 & \textbf{25.27} & 46.69 & \textbf{60.89} & 10.52 & \textbf{14.66} & 8.06 & \textbf{19.71} \\
Llama3 13B & 50.21 & \textbf{73.97} & 18.61 & \textbf{21.76} & 15.42 & \textbf{17.61} & 42.12 & \textbf{47.26} & 5.84 & \textbf{9.13} & 4.01 & \textbf{6.16} \\ 

\bottomrule
\end{tabular}
}
\caption{Performance of LLMs in \emph{Text-to-Triple} \xinbang{Pipeline}. \emph{Text} input is the document, and \emph{Triple} input is randomly shuffled linearized triples.}
\label{tab1}
\end{table}

\paragraph{Results Analysis}
The data in Table~\ref{tab1} indicate that, compared to NL text, LLMs achieve significant improvements when using linearized triples, a finding consistent with those from \emph{Triple-to-Text}. Additionally, we observe that for multi-hop, fact-intensive questions, knowledge prompts in text format may hinder LLMs from providing better answers. When addressing multi-hop questions, text format inputs require LLMs to perform reasoning across sentences or paragraphs, where unrelated information can adversely affect LLM performance. For instance, fluent NL texts may introduce noise (such as function words and determiners), hindering LLMs' ability to recognize the core knowledge it should consider. LLMs do not struggle with understanding unordered linearized triples; on the contrary, this straightforward prompt type is more conducive to answering multi-hop questions.
%
% \xinbang{In the domain-specific KG, LLMs' performance declines compared to the common KG, Wikidata. This indicates that LLMs have a weaker grasp of domain-specific knowledge than general world knowledge. Even when external knowledge is provided, enabling LLMs to comprehend and utilize domain-specific information to answer questions remains a significant challenge.}

\devin{In the domain-specific KG, compared to the general KG (Wikidata), the performance of LLMs shows a noticeable decline. This decline can be attributed to the fact that specialized domain knowledge is often not included in the pretraining corpus of LLMs. Consequently, LLMs lack the relevant expertise in specialized fields, which may result in a weaker grasp of domain-specific knowledge compared to general world knowledge. Notably, even when external knowledge is provided, enabling LLMs to understand and effectively utilize domain-specific information to answer questions remains a significant challenge.}

Based on the complementary experiments of \emph{Triple-to-Text} and \emph{Text-to-Triple}, we draw an important conclusion: LLMs tend to prefer unordered linearized triples as the input format of KG.

\subsection{Attention Level Evaluation of LLM's Understanding of KG}
\label{sec:3.2}

This section explores how LLMs utilize KG information in different input formats, further revealing their ability to understand KGs. Specifically, we observe the attention distribution of LLMs towards answers in different KG input formats. This distribution illustrates the information's prioritized degree, thereby assisting us in analyzing which input format of KG is more beneficial for prompting LLMs. We introduce our evaluation method and then employ a quantitative analysis based on two datasets.

\begin{figure}[t]
\centering
\includegraphics[width=\columnwidth]{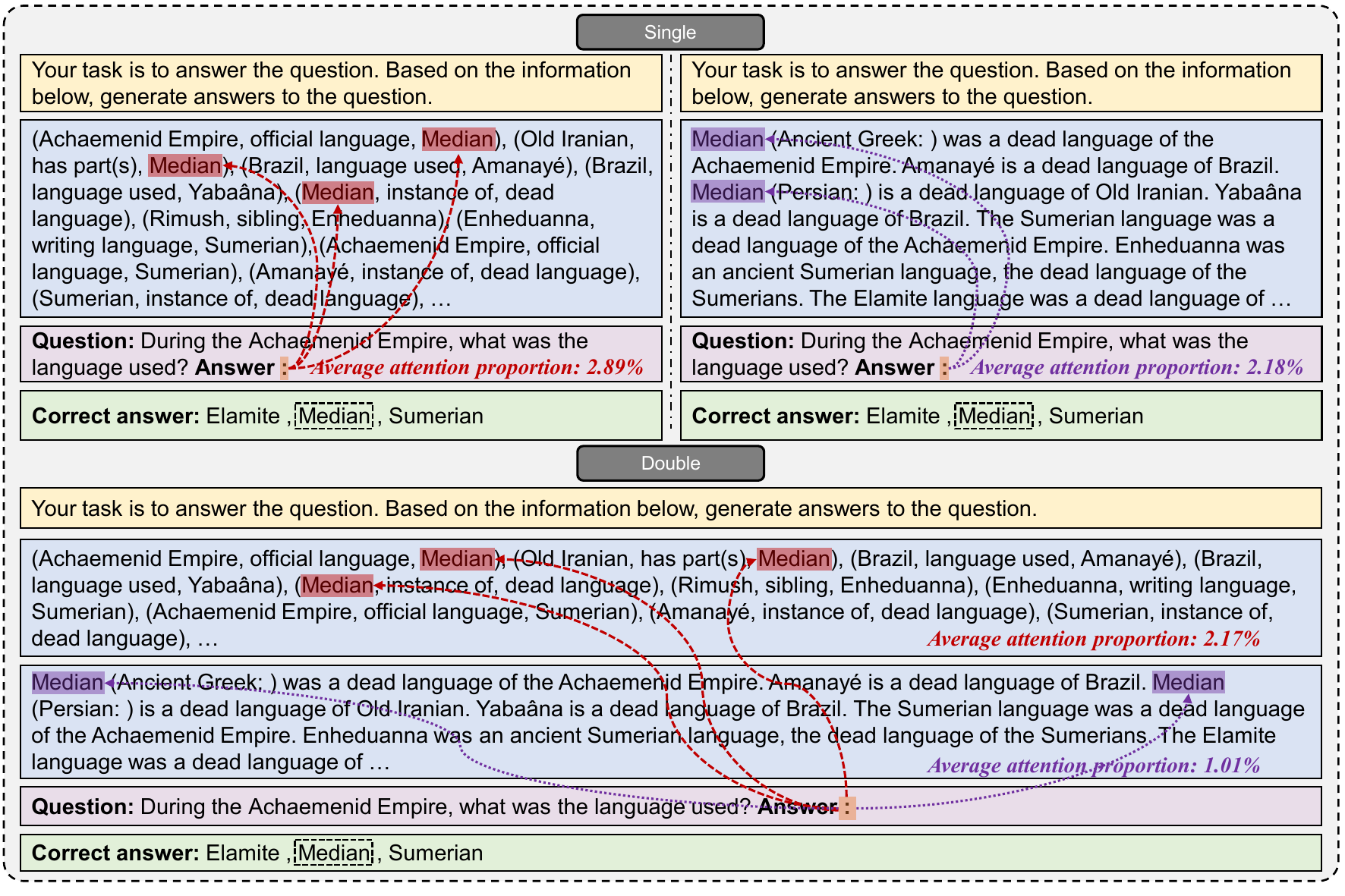}
\caption{We examine the average attention proportion between the predicted labels (indicated by a colon ":") and the answer words (e.g., "Median"). Whether in the Single mode (providing knowledge in one format) or the Double mode (providing knowledge in both formats simultaneously), the LLM consistently pays more attention to the answers in the linearized triple format.}
\label{fig5}
\end{figure}

\subsubsection{Attention Level Evaluation Method}

As shown in Figure~\ref{fig5}, the input format of knowledge is divided into unordered linearized triples $g_u$ and NL text $g_f$. We employ two fair comparison modes for knowledge prompts: (1) Single: providing the LLM with only one format of knowledge (either $g_u$ or $g_f$); (2) Double: providing the LLM with both formats of knowledge simultaneously ($g_u$ and $g_f$).

% The input's last token, $p$, is the model's prediction label~\cite{wang2023label}. 
\devin{During inference, the last token of the input, $p$, is used to trigger the model's predicted label~\cite{wang2023label}.}
% We focus on the attention proportion of $p$ towards the answers within the distribution, which reflects the concerning of LLM to the answers in the provided knowledge. 
\devin{We examine the proportion of $p$'s attention directed toward the tokens within the input that belong to the answer.
This reflects the degree to which the LLM focuses on the answer within the provided knowledge.}
A higher proportion suggests a greater likelihood that the model will include the answer in its generated response. Considering that the number of answer occurrences may vary across different knowledge formats, we calculate the average attention proportion for each answer to ensure fairness. Therefore, we calculate the average attention proportion between the last token (the colon ":" ) and the answer word (for instance, "Median") as follows:

\begin{equation}
    \overline{Att} = \frac{1}{n}\sum_{i=1}^{n}Att_{ans_{i}, p}
\end{equation}

Here, \(Att_{ans_{i}, p}\) represents the attention proportion between all tokens of answer \(i\) in the prompt and the prediction label \(p\), and \(n\) denotes the total occurrences of answer tokens in the prompt.

\subsubsection{Experimental Setup}
\paragraph{Datasets and Metric} 
Our study selects two datasets from the \emph{Triple-to-Text} experiment: LC-QuAD 2.0 and KQAPro. We utilize subgraphs with parameters \emph{expanded\_nodes=2} and \emph{depth=2}. The input modes are unordered linearized triples \(g_u\) and fluent NL text \(g_f\). For a dataset \(D\) containing \(m\) questions, the attention proportion for question \(q_k\) (\(k=1, \ldots, m\)) is denoted as \(\overline{Att}_k\). Consequently, the attention score of the LLM for the answers within the dataset \(D\) is calculated as:

\begin{equation}
Att_D = \frac{1}{m}\sum_{k=1}^{m}\overline{Att}_k
\end{equation}

\paragraph{LLMs} We use the open-source Vicuna 7B and 13B models to examine how LLMs of different scales understand subgraphs with two input formats. In the final layer of the model, we sum the attention between each token and the prediction token $p$ across all heads~\cite{wang2023label}. Then, we normalize all values to obtain each token's attention proportion of $p$.

\begin{table}[htbp]
\centering
\resizebox{0.75\columnwidth}{!}{%
\small
\begin{tabular}{lcccccccc}
\toprule
\multirow{3}{*}{} & \multicolumn{4}{c}{Vicuna 7B} & \multicolumn{4}{c}{Vicuna 13B} \\ 

\cmidrule(r){2-5} \cmidrule(r){6-9}

 & \multicolumn{2}{c}{KQAPro} & \multicolumn{2}{c}{LC-QuAD 2.0} & \multicolumn{2}{c}{KQAPro} & \multicolumn{2}{c}{LC-QuAD 2.0} \\ 
 
\cmidrule(r){2-5} \cmidrule(r){6-9}

 & $g_f$ & $g_u$ & $g_f$ & $g_u$ & $g_f$ & $g_u$ & $g_f$ & $g_u$ \\ 

\cmidrule(r){1-1} \cmidrule(r){2-3} \cmidrule(r){4-5} \cmidrule(r){6-7} \cmidrule(r){8-9}

Single & 2.46 & \textbf{3.80} & 2.08 & \textbf{2.67} & 2.78 & \textbf{4.69} & 2.10 & \textbf{3.78} \\
Double & 1.12 & \textbf{1.31} & 0.89 & \textbf{2.38} & 1.62 & \textbf{2.72} & 1.40 & \textbf{2.57} \\ \bottomrule
\end{tabular}%
}
\caption{The average attention proportion of LLM to the answer tokens in the prompt.}
\label{tab2}
\end{table}

\paragraph{Results Analysis}

The experimental results, as shown in Table~\ref{tab2}, indicate that whether the two types of knowledge are provided simultaneously or separately, the model consistently shows a higher attention ratio to linearized triples. 
%
% This suggests that LLMs can accommodate less fluent knowledge prompts.
%
This suggests that LLMs can adapt to knowledge prompts that are less similar to the NL input format.
Furthermore, when addressing fact-intensive questions, the LLM can identify the critical information and retrieve answers from linearized triple prompts. This explains why the LLM demonstrates greater interest in unordered linearized triples than NL text.

\section{Comparing LLM's Performance on Different KG Prompt Strategies}

% Based on the findings in Section 4, we observe that unordered linearized triples used as prompts yield more benefits for LLMs than fluent NL text.
%
Based on the findings in Section~\ref{sec:3}, we observe that unordered linearized triples used as prompts yield more benefits for LLMs than fluent NL text.
However, these triples are randomly ordered in prompts and lack a specific organization. Therefore, discussing the organization of linearized triples is essential for designing LLM-friendly prompts.
In this section, we design various prompt strategies for linearized triples based on the \emph{question relevance score}. We discover that different versions of LLMs exhibit distinct preferences for these strategies, highlighting the need to explore universal prompting techniques.

\subsection{Question Relevance Score}

\subsubsection{Cross-encoder Based Scorer}
To establish the criteria for sorting triples, we use the relevance score between the triples and the question. We construct a cross-encoder based on the BERT-base~\cite{devlin2018bert} model, similar to those described in~\cite{logeswaran2019zero} and~\cite{humeau2019poly}. The input $\tau_{q, t}$ consists of the concatenation of the question and the triple, enabling the model to develop deep cross-attention between them. Formally, we denote the joint embedding of the question and triple as \( v_{q, t} \):

\begin{equation}
v_{q, t} = \text{red}(\text{BERT}_{\text{cross}}(\tau_{q, t}))
\end{equation}

Here, \( \tau_{q, t} \) represents the input representation of the question and triple, \(\text{BERT}_{\text{cross}}\) is the BERT-based cross-encoder, and the function \(\text{red}(.)\) reduces the sequence of vectors produced by the encoder into one vector. 
Based on the experiments in~\cite{humeau2019poly}, we select the \(\text{red}(.)\) function as the last layer of the output of the [CLS] token. To score candidate entities, we apply a linear layer \( W \) to the embedding \( v_{q, t} \):

\begin{equation}
s_{\text{cross}}(q, t) = v_{q, t}W
\end{equation}

\subsubsection{Experimental Setup}
The scorer is based on the BERT-base model. We divide all the triples and questions from the \emph{Text-to-Triple} dataset into the training set and a test set with an 8:2 ratio. Triples in the reasoning path linked to the question are labelled positive examples; otherwise, they are designated negative examples. For the cross-encoder, the batch size is 50; we experiment with initial learning rates of \{5e-4, 2e-5, 5e-5, 2e-5\}, and the learning rate decays every 3 epochs. We set the multiplicative factor, gamma, to update the learning rate to 0.2. Upon training, the model exhibits an accuracy of 98.89$\%$ in determining whether triples are related to the question, i.e., whether they are part of the core reasoning path.

\subsection{Prompt Strategies}
\label{sec:4.2}

We perform the following operations on these triples using \(s_{\text{cross}}(q, t)\): 
(1) Grouping: By setting thresholds at 0.3 and 0.8, we categorize the triples into high, medium, and low relevance groups relative to the question.
(2) Ranking: We sort the candidate triples in descending order according to \(s_{\text{cross}}(q, t)\).
(3) Scoring: We append \(s_{\text{cross}}(q, t)\) to each triple to indicate the relevance between the question and the triples to the LLM.

Finally, we input the triples and questions into the LLM to generate the answers. Table~\ref{tab3} illustrates three prompting strategies for the question: \emph{What business structure did Frank Gehry design?} For grouping, we require the LLM to focus on higher relevant triples, thereby aiding in narrowing the search scope. For ranking, we expect the LLM to prioritize the foremost information in a sequence of triples. For scoring, we hope that the score will assist LLMs in retrieving relevant triples.

\subsection{Datasets and Metric}
We utilize all datasets from the \emph{Triple-to-Text} experiment. We still employ subgraphs with parameters set to \emph{expanded\_nodes=2} and \emph{depth=2}. The originally unordered linearized triples are combined using three prompt strategies. 
\xinbang{The evaluation metrics follow the experimental setup in Section~\ref{ssec:3.1.1}.}

\begin{table}[t]
\centering
\resizebox{\columnwidth}{!}{%
\small
\begin{tabular}{p{0.1\linewidth} p{0.9\linewidth}}
\toprule
Strategies & \multicolumn{1}{c}{Prompt} \\ 
\midrule
Grouping & You are a QA assistant. For question: \emph{What business structure did Frank Gehry design?} Refer to the following knowledge to response.\newline \textbf{Here are some triples that are highly relevant to the question:} (DZ Bank building, architect, Frank Gehry), (Gehry Tower, instance of, office building), ... \textbf{Here are some triples that are likely relevant to the question:} (IAC Building, architect, Frank Gehry), (Gehry Tower, architect, Frank Gehry) ... \textbf{Here are some triples that are less relevant to the question:} (Toledo Museum of Art, architect, Frank Gehry), (Vlado Miluni, notable work, Dancing House), ... \newline Answer:\\
\midrule
Ranking & You are a QA assistant. For question: \emph{What business structure did Frank Gehry design?} Refer to the following knowledge to response.\newline \textbf{These triples are sorted from high to low according to their relevance score to the question:} (DZ Bank building, architect, Frank Gehry), (Dancing House, instance of, office building),(Gehry Tower, architect, Frank Gehry), (Dancing House, architect, Frank Gehry), (IAC Building, instance of, office building), ... \newline Answer:\\
\midrule
Scoring & You are a QA assistant. For question: \emph{What business structure did Frank Gehry design?} Refer to the following knowledge to response.\newline \textbf{Each triple is followed by a relevance score to the question, which helps in solving the question:} (DZ Bank building, architect, Frank Gehry) | 0.9981, (Toledo Museum of Art, architect, Frank Gehry) | 0.0019, Gehry Tower, instance of, office building) | 0.998, (Vlado Miluni, notable work, Dancing House) | 0.0023... \newline Answer:\\ 
\bottomrule
\end{tabular}%
}
\caption{Different prompt strategies for triples based on relevance scores.}
\label{tab3}
\end{table}

\subsection{Results Analysis}

The experimental results in Table~\ref{tab4} indicate that, compared to the unordered linearized triplet input \(g_u\) with parameters set as \emph{expanded\_nodes=2} and \emph{depth=2}, a well-designed prompt strategy can enhance model performance. However, the ranking operation does not necessarily improve the performance of all LLMs; they have inconsistent preferences for different prompt strategies. For instance, ChatGPT favors the prompt method incorporating relevance scores, while the Vicuna series prefers ranking strategies. This discrepancy may be attributed to variations in the training data and inherent structure of the respective models. This finding suggests that when designing a prompt method, the universal applicability of a given prompt strategy across multiple LLMs should be considered.

\begin{table}[htbp]
\centering
\resizebox{\columnwidth}{!}{%
\setlength{\tabcolsep}{1.3pt}
\begin{tabular}{lcccccccccccc}
\toprule
\multicolumn{1}{c}{} & \multicolumn{4}{c}{ChatGPT} & \multicolumn{4}{c}{Vicuna 7B} & \multicolumn{4}{c}{Vicuna 13B} \\ 

\cmidrule(r){2-5} \cmidrule(r){6-9} \cmidrule(r){10-13}

Datasets & $g_u$ & Grouping & Ranking & Scoring & $g_u$ & Grouping & Ranking & Scoring & $g_u$ & Grouping & Ranking & Scoring \\ 

\cmidrule(r){1-1} \cmidrule(r){2-5} \cmidrule(r){6-9} \cmidrule(r){10-13}

QALD-7 & 82.56 & 84.11 & 84.11 & \textbf{84.11} & 65.70 & 64.84 & \textbf{66.54} & 53.64 & 77.00 & 75.52 & \textbf{77.81} & 72.40 \\
LC-QuAD 2.0 & 50.60 & 48.71 & 50.01 & \textbf{52.48} & 32.84 & 33.49 & \textbf{35.72} & 26.14 & 42.98 & 45.10 & \textbf{45.13} & 42.57 \\
KQAPro & 54.03 & 50.25 & 52.29 & \textbf{54.03} & 27.63 & 27.74 & \textbf{31.32} & 24.92 & 32.75 & 36.05 & \textbf{37.64} & 35.12 \\ 
\bottomrule
\end{tabular}%
}
\caption{Distinct models exhibit unique preferences towards various prompting methods.}
\label{tab4}
\end{table}

\section{Evaluating LLM's Robustness to Noisy or Incomplete KG}
In real-world scenarios, the subgraphs provided to LLMs often contain noise or are incomplete. Therefore, understanding the robustness of LLMs when handling noisy or incomplete subgraphs is crucial. In this section, we gradually corrupt the subgraphs to observe the performance of LLMs in linearized triple input formats.

\subsection{The Generation of Noisy or Incomplete Subgraphs}

\begin{figure}[htbp]
\centering
\includegraphics[width=\columnwidth]{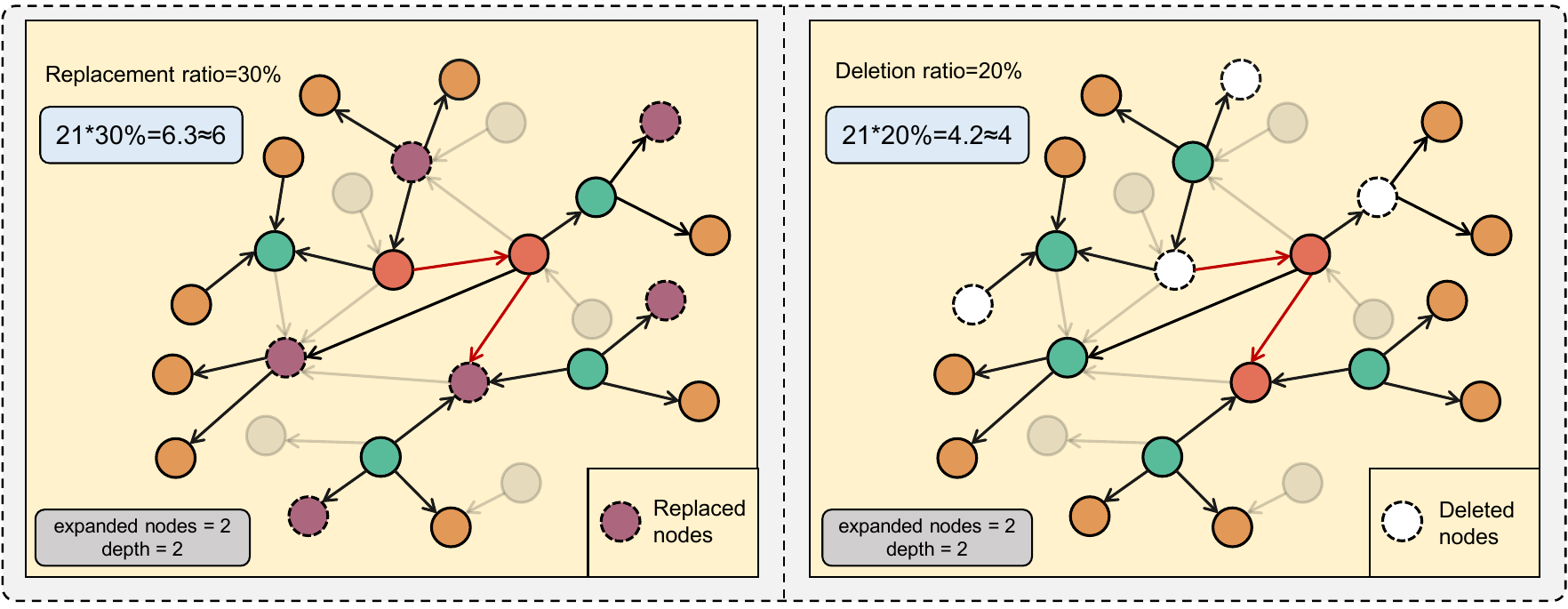}
\caption{We calculate the number of nodes for random replacement and deletion based on the ratio. The nodes used for replacement are randomly sampled from unrelated nodes in the KG.}
\label{fig6}
\end{figure}

As illustrated in Figure~\ref{fig6}, we systematically corrupt the subgraph in two approaches: 1) Nodes are proportionally replaced with random irrelevant KG nodes. 2) Nodes are proportionally deleted randomly.

In the replacement operation, unrelated nodes replace parts of gloden nodes to generate false fact information, denoted as a noisy subgraph. The deletion operation simulates an incomplete KG scenario. We randomly replace and delete nodes according to specified percentages. The replacement and deletion ratio ranges from 10$\%$ to 90$\%$. 

\subsection{Datasets and Metric}
Based on all datasets from the \emph{Triple-to-Text} experiments, we utilize subgraphs with parameters set to \emph{expanded\_nodes=2} and \emph{depth=2}, where nodes are randomly replaced or deleted. The evaluation metrics adhere to those used in the \xinbang{Section~\ref{ssec:3.1.1}}. We aim to observe the robustness of different LLMs when the proportion of replacements and deletions increases.

\subsection{Results Analysis}

As shown in Table~\ref{tab5}, across three datasets, ChatGPT shows greater performance degradation on replacement and deletion compared to the smaller parameter models, Vicuna 7B and 13B.
We have two preliminary findings: 1) The random replacement of nodes in KG has a more significant negative impact on the inference performance of LLM compared to the random deletion. \emph{Incorrect facts are more likely to result in erroneous model outputs.} 2) Despite larger models demonstrating superior answering performance, they exhibit a more significant performance loss when facing random replacement and deletion of the subgraph. \emph{There exists an inverse proportionality between a model's robustness and its size.}

\begin{table}[htbp]
\setlength{\tabcolsep}{1.5pt}
\resizebox{\columnwidth}{!}{%
\begin{tabular}{ccccccccccccccccccc}
\toprule
 & \multicolumn{6}{c}{QALD-7} & \multicolumn{6}{c}{LC-QuAD 2.0} & \multicolumn{6}{c}{KQAPro} \\ \cmidrule(r){2-7} \cmidrule(r){8-13} \cmidrule(r{0.1em}){14-19}
\multicolumn{1}{l}{} & \multicolumn{2}{c}{ChatGPT} & \multicolumn{2}{c}{Vicuna 7b} & \multicolumn{2}{c}{Vicuna 13b} & \multicolumn{2}{c}{ChatGPT} & \multicolumn{2}{c}{Vicuna 7b} & \multicolumn{2}{c}{Vicuna 13b} & \multicolumn{2}{c}{ChatGPT} & \multicolumn{2}{c}{Vicuna 7b} & \multicolumn{2}{c}{Vicuna 13b} \\ 

\cmidrule(r{0.1em}){1-1} \cmidrule(r){2-7} \cmidrule(r){8-13} \cmidrule(r{0.1em}){14-19}

Ratio & Delete & Replace & Delete & Replace & Delete & Replace & Delete & Replace & Delete & Replace & Delete & Replace & Delete & Replace & Delete & Replace & Delete & Replace \\ 

\cmidrule(r{0.1em}){1-1} \cmidrule(r){2-7} \cmidrule(r){8-13} \cmidrule(r{0.1em}){14-19}

0\% & 82.56 & 82.56 & 65.70 & 65.70 & 77.00 & 77.00 & 50.60 & 50.60 & 32.84 & 32.84 & 42.98 & 42.98 & 54.03 & 54.03 & 27.63 & 27.63 & 32.75 & 32.75 \\
10\% & 82.08 & 80.62 & 64.44 & 58.89 & 72.85 & 78.21 & 47.84 & 47.74 & 31.07 & 32.20 & 41.66 & 40.55 & 51.57 & 52.04 & 26.12 & 26.23 & 31.32 & 31.24 \\
20\% & 82.08 & 80.62 & 70.00 & 58.65 & 75.51 & 81.16 & 46.23 & 46.02 & 31.83 & 29.59 & 39.43 & 39.43 & 49.66 & 49.91 & 26.09 & 26.17 & 30.96 & 31.07 \\
30\% & 80.39 & 79.90 & 62.95 & 59.08 & 74.59 & 78.70 & 44.18 & 43.60 & 29.51 & 28.10 & 38.57 & 38.85 & 46.95 & 46.87 & 23.52 & 24.58 & 28.94 & 29.81 \\
40\% & 80.39 & 79.90 & 55.27 & 57.68 & 77.00 & 73.84 & 42.49 & 42.41 & 27.83 & 27.47 & 36.12 & 36.83 & 43.82 & 44.10 & 23.60 & 23.94 & 27.93 & 28.19 \\
50\% & 80.39 & 80.39 & 66.86 & 59.66 & 76.52 & 74.35 & 40.78 & 40.28 & 27.78 & 26.03 & 34.57 & 35.09 & 41.02 & 41.33 & 21.95 & 21.87 & 26.26 & 27.88 \\
60\% & 80.39 & 80.39 & 63.24 & 61.30 & 75.31 & 78.21 & 36.25 & 35.99 & 24.50 & 22.56 & 33.50 & 32.94 & 37.28 & 37.08 & 21.37 & 19.16 & 25.53 & 25.36 \\
70\% & 75.56 & 75.56 & 63.00 & 48.74 & 73.62 & 68.07 & 31.87 & 31.72 & 21.82 & 19.44 & 30.54 & 27.36 & 32.33 & 32.02 & 18.06 & 16.69 & 23.01 & 23.38 \\
80\% & 65.85 & 63.91 & 50.00 & 47.25 & 65.17 & 66.14 & 26.28 & 25.82 & 18.54 & 16.50 & 25.56 & 23.79 & 27.21 & 27.46 & 14.57 & 13.49 & 20.41 & 20.25 \\
90\% & 62.42 & 54.30 & 46.64 & 45.02 & 57.51 & 56.23 & 19.16 & 17.64 & 13.21 & 11.98 & 18.48 & 17.51 & 19.91 & 19.77 & 10.49 & 9.82 & 14.15 & 14.90 \\ 

\cmidrule(r{0.1em}){1-1} \cmidrule(r){2-7} \cmidrule(r){8-13} \cmidrule(r{0.1em}){14-19}

$\downarrow$ & \textcolor[rgb]{0.25, 0.5, 0.75}{20.14} & \textcolor[rgb]{0.25, 0.5, 0.75}{28.26} & \textcolor[rgb]{0.25, 0.75, 0.25}{19.06} & \textcolor[rgb]{0.25, 0.75, 0.25}{20.68} & \textcolor[rgb]{0.75, 0.25, 0.5}{19.49} & \textcolor[rgb]{0.75, 0.25, 0.5}{20.77} & \textcolor[rgb]{0.25, 0.5, 0.75}{31.44} & \textcolor[rgb]{0.25, 0.5, 0.75}{32.96} & \textcolor[rgb]{0.25, 0.75, 0.25}{19.63} & \textcolor[rgb]{0.25, 0.75, 0.25}{20.86} & \textcolor[rgb]{0.75, 0.25, 0.5}{24.50} & \textcolor[rgb]{0.75, 0.25, 0.5}{25.47} & \textcolor[rgb]{0.25, 0.5, 0.75}{34.12} & \textcolor[rgb]{0.25, 0.5, 0.75}{34.26} & \textcolor[rgb]{0.25, 0.75, 0.25}{17.14} & \textcolor[rgb]{0.25, 0.75, 0.25}{17.81} & \textcolor[rgb]{0.75, 0.25, 0.5}{18.60} & \textcolor[rgb]{0.75, 0.25, 0.5}{17.85} \\ 
\bottomrule

\end{tabular}
}
\caption{Randomly delete and replace nodes in the subgraph. $\downarrow$ quantifies the discrepancy between the model's peak performance and its poorest performance.}
\label{tab5}
\end{table}

\section{Conclusion}

In this study, we empirically study various input formats of KG injected into LLMs, yielding fundamental insights. For fact-intensive questions, linearized triples prompt LLMs more effectively than NL text. Additionally, we designed several simple prompt strategies to observe LLMs' preference for the organization of triples, and the experiment results show that different series of LLMs do not necessarily rely on order. We also evaluate the robustness of LLMs to noisy or incomplete subgraphs, discovering that larger models are more susceptible to these issues.
%
% The experiments suggest that when designing prompt strategies for larger LLMs, the impact of subgraph quality should also be considered.
%
The experiments suggest that when designing prompt strategies for larger LLMs, the impact of noise in the subgraph should also be considered.
In summary, our findings offer valuable guidance for refining KG-related prompt strategies and underscore the importance of linearized triple knowledge in prompting LLMs.

\section*{Acknowledgments}
This work is supported by the Natural Science Foundation of China (Grant No. U21A20488). We thank the Big Data Computing Center of Southeast University for providing the facility support on the numerical calculations in this paper.

\bibliographystyle{elsarticle-num}
\bibliography{ref}

\end{document}